\documentclass[10pt,twocolumn,letterpaper]{article}

\usepackage{iccv}              

\usepackage{lipsum}
\usepackage{bbm}
\usepackage{xspace}
\usepackage{graphicx}
\usepackage{tabu}
\usepackage{multirow}
\usepackage{pifont}
\usepackage{amsmath}
\usepackage{listings}

\newcommand{\red}[1]{{\color{red}#1}}

\definecolor{iccvblue}{rgb}{0.21,0.49,0.74}
\usepackage[pagebackref,breaklinks,colorlinks,allcolors=iccvblue]{hyperref}

\def\paperID{382} 
\def\confName{ICCV}
\def\confYear{2025}

\title{Inter2Former: Dynamic Hybrid Attention for  \\Efficient High-Precision Interactive Segmentation}

\author{%
    You Huang, Lichao Chen, Jiayi Ji, Liujuan Cao, Shengchuan Zhang\thanks{Corresponding author}, Rongrong Ji \\
    Key Laboratory of Multimedia Trusted Perception and Efficient Computing, \\
    Ministry of Education of China, Xiamen University \\
    {\tt\small youhuang0607@gmail.com, fpg2012@foxmail.com, jjyxmu@gmail.com}\\
    {\tt\small caoliujuan@xmu.edu.cn, zsc\_2016@xmu.edu.cn, rrji@xmu.edu.cn}\\
}

\begin{document}
\maketitle
\begin{abstract}
Interactive segmentation (IS) improves annotation efficiency by segmenting target regions from user prompts, with widespread applications in real-world scenarios. Current approaches face a critical trade-off: dense-token methods achieve superior accuracy and detail preservation but suffer from prohibitively slow processing on CPU devices, while the Segment Anything Model (SAM) advances the field with sparse prompt tokens for fast inference but compromises segmentation quality. In this paper, we propose Inter2Former to address this challenge by optimizing computation allocation in dense-token processing, which introduces four key enhancements. First, we propose Dynamic Prompt Embedding (DPE) that adaptively processes only regions of interest while avoiding additional overhead from background tokens. Second, we introduce Dynamic Hybrid Attention (DHA), which leverages previous segmentation masks to route tokens through either full attention ($O(N^2)$) for boundary regions or our proposed efficient BSQ attention ($O(N)$) for non-boundary regions. Third, we develop Hybrid Mixture of Experts (HMoE), which applies similar adaptive computation strategies in FFN modules with CPU-optimized parallel processing. Finally, we present Dynamic Local Upsampling (DLU), a reverse operation of DPE, which localizes objects with a lightweight MLP and performs fine-grained upsampling only in detected regions. 
Experimental results on high-precision IS benchmarks demonstrate that Inter2Former achieves SOTA performance with high efficiency on CPU devices. 
\end{abstract}    
\section{Introduction}
Interactive segmentation  (IS)~\cite{kirillov2023segany,QinLiu2022SimpleClickII,XiChen2022FocalClickTP,huang2023interformer} greatly enhances the image segmentation annotation process by segmenting regions of interest from a few annotator prompts, such as clicks~\cite{huang2023interformer}, bounding boxes~\cite{kirillov2023segany} or coarse masks~\cite{KonstantinSofiiuk2021RevivingIT}. These works expand real-world applications of image segmentation, \eg medical imaging~\cite{SaberMirzaeeBafti2021ACS}, industrial defect detection~\cite{bergmann2019mvtec} and autonomous driving~\cite{HolgerCaesar2019nuScenesAM}. Recently, the Segment Anything Model (SAM)~\cite{kirillov2023segany} has become a milestone in IS, excelling in real-time, high-quality segmentation, particularly in mainstream click-based IS.

\begin{figure}[t]
  \centering
  \includegraphics[width=0.45\textwidth]{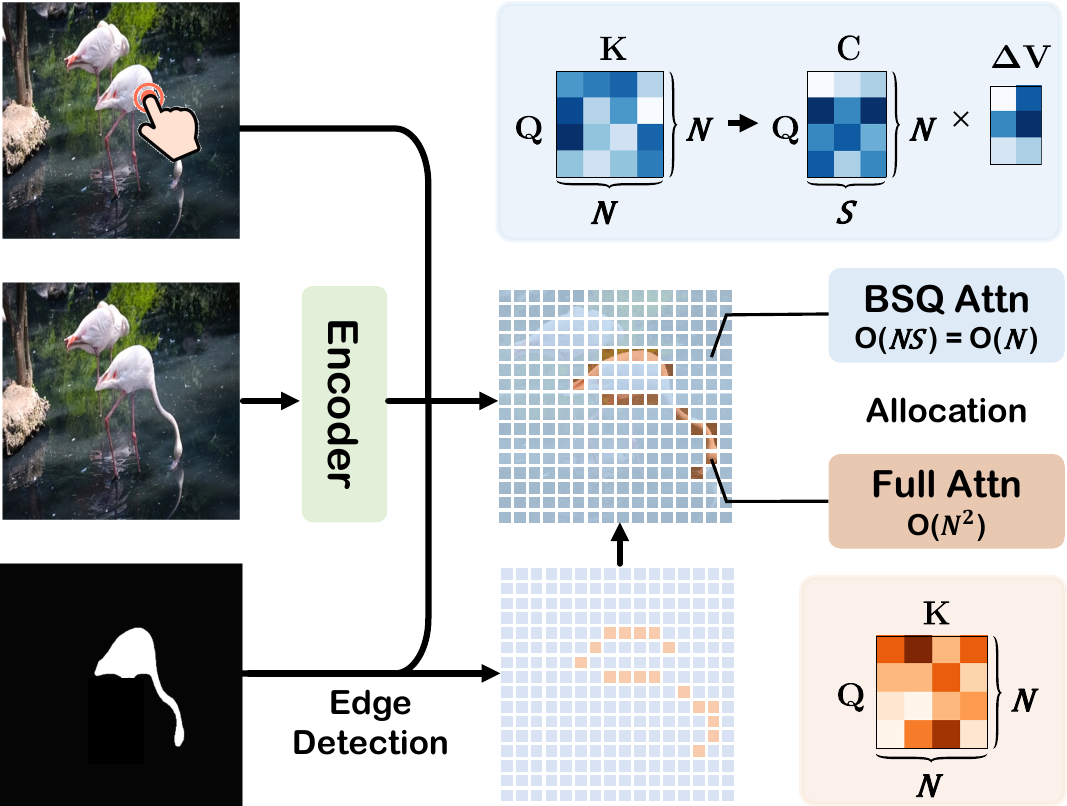}
  \caption{Overview of Dynamic Hybrid Attention (DHA). During interactive segmentation, the model receives user clicks (top) and the previous segmentation mask (bottom). Our DHA leverages mask boundaries to route image tokens through either Full Attention with $O(N^2)$ complexity for boundary regions, or our proposed BSQ Attention with $O(NS) = O(N)$ complexity for non-boundary regions, optimizing computation while preserving segmentation quality.}
  \label{fig:teaser}
  \vspace{-1em}
\end{figure}

SAM~\cite{kirillov2023segany} and InterFormer~\cite{huang2023interformer} simultaneously introduce a similar two-stage pipeline for this task: a preprocessing stage that encodes images into tokens using an encoder, followed by an interaction stage where a decoder processes these tokens along with user prompts to generate segmentation masks. While sharing similar encoders, these models differ significantly in their decoder design. InterFormer~\cite{huang2023interformer} converts clicks into dense prompt tokens to enhance spatial awareness and achieve superior segmentation accuracy. However, this approach incurs high computational costs, making it impractically slow on CPU devices, a significant limitation for large-scale crowdsourced annotation with limited GPU resources. In contrast, SAM~\cite{kirillov2023segany} utilizes sparse prompt tokens for efficient cross-attention and faster inference, but sacrifices spatial awareness and boundary precision as a result. Subsequently, SegNext~\cite{liu2024rethinking} improves SAM's decoder by incorporating dense prompt tokens, which enhances accuracy but further increases computational demands. HRSAM~\cite{huang2024hrsam} focuses on upgrading SAM's encoder while maintaining its efficient but less accurate sparse-token decoder. Despite these efforts, achieving high-precision interactive segmentation under computational constraints remains challenging~\cite{liu2024rethinking,huang2024hrsam}, especially for high-resolution images where the performance-efficiency trade-off becomes more severe.

To address the aforementioned challenges, we investigate an approach built upon dense prompt tokens~\cite{huang2023interformer,liu2024rethinking} while maintaining efficient inference capabilities. We observe that the inefficiency of dense prompt tokens stems from suboptimal resource allocation in the IS process. Specifically, existing models uniformly allocate computation across all tokens during the entire IS process. However, the main object region is typically determined within the first few clicks and most subsequent user clicks focus on refining object boundaries. Computational resources should therefore be prioritized for boundary regions. Thus, this uniform allocation wastes computations on already determined main object regions while providing insufficient computation to critical boundary regions, leading to suboptimal precision and efficiency. Furthermore, each step's segmentation result in the IS process contains boundary cues, but existing models simply incorporate the segmentation result as an additional feature~\cite{KonstantinSofiiuk2021RevivingIT,QinLiu2022SimpleClickII,kirillov2023segany,huang2023interformer} to refine the next interaction, failing to fully utilize the boundary information for computational optimization.

In this paper, we introduce Inter2Former\footnote{Code is available at \url{https://github.com/YouHuang67/inter2former}.}, optimizing computation allocation for efficient high-precision IS. First, we propose Dynamic Prompt Embedding (DPE) that processes only regions of interest while avoiding computational overhead of background tokens through dynamic region cropping. Second, we propose Dynamic Hybrid Attention (DHA) to allocate computational resources differentially between dynamically detected boundary and non-boundary tokens. In DHA (Figure~\ref{fig:teaser}), boundary tokens are processed through conventional Full Attention (FA)~\cite{AshishVaswani2017AttentionIA}, while non-boundary tokens utilize our lightweight BSQ Attention (BSQA). Our BSQA, inspired by~\cite{lingle2023transformer}, applies Binary Spherical Quantization~\cite{zhao2024image} to compress key-value pairs, reducing complexity from $O(N^2)$ to $O(N)$ for non-boundary tokens. Third, we propose Hybrid Mixture of Experts (HMoE) in FFN modules to dynamically route boundary and non-boundary tokens to either MoE or conventional FFN for optimized computation allocation. Besides, we optimize MoE computation on CPUs by rearranging tokens to enhance low-level matrix computation and reduce latency. Fourth, we propose Dynamic Local Upsampling (DLU) to speed up mask prediction, functioning as an inverse operation to DPE. DLU first employs a lightweight MLP and Canny operator to localize regions of interest, then performs upsampling exclusively within these regions for fine-grained segmentation at minimal computational cost.

We evaluate Inter2Former on high-precision interactive segmentation benchmarks~\cite{liu2024rethinking,huang2024hrsam,FedericoPerazzi2016ABD,sam_hq}. Our method achieves SOTA performance with slight additional latency over sparse-token models~\cite{kirillov2023segany,huang2024hrsam}. The CPU-optimized HMoE reduces inference time by $56$-$85\%$ over vanilla MoE implementations.

Our main contributions are as follows:
\begin{itemize}
    \item We propose DHA, which assigns tokens to Full Attention (FA) or our novel BSQ Attention (BSQA) based on prior segmentation results, optimizing computation allocation.  
    \item We propose HMoE for better computation allocation, with optimized parallel MoE computation on CPUs through token rearrangement, reducing inference latency by $56$-$85\%$ over vanilla MoE implementations.
    \item We propose DPE and DLU for efficient prompt encoding and mask prediction. DPE dynamically crops regions of interest, while DLU performs the inverse operation by using a lightweight MLP to identify object regions for selective upsampling and fine-grained segmentation.
    \item Inter2Former achieves SOTA while maintaining high efficiency on CPU devices.  
\end{itemize}

\begin{figure*}[t]
  \centering
  \includegraphics[width=1\textwidth]{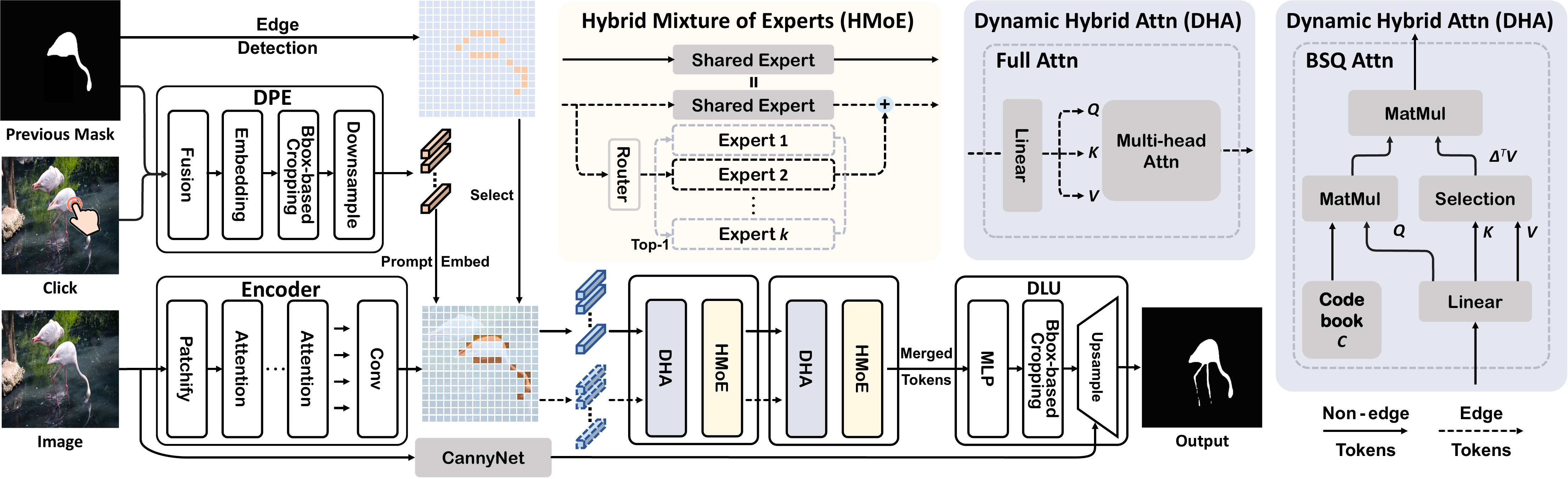}
  \vspace{-1em}
  \caption{Overview of Inter2Former. Given user clicks and previous mask as interactive prompts, our Inter2Former first employ DPE to fuse and transform the prompts into prompt embeddings that are then added to image tokens extracted by the encoder. Based on edge detection from the previous mask, the prompt-fused tokens are divided into edge tokens and non-edge tokens for efficient processing. These tokens undergo DHA and HMoE layers. In DHA, edge tokens are processed by Full Attention while non-edge tokzens utilize our proposed BSQA. In HMoE, edge tokens are processed by both multiple routed experts and a shared expert, followed by summation of their outputs, while non-edge tokens only pass through the shared expert. Finally, the merged tokens are upsampled through DLU to generate the output mask.}
  \label{fig:overview}
\end{figure*}

\section{Related Work}
\noindent\textbf{Interactive Segmentation.}
Interactive segmentation is first integrated with deep learning in DIOS~\cite{NingXu2016DeepIO}, which introduces deep neural networks to this field and establishes the standard training and evaluation protocol for mainstream click-based interaction. Subsequent research advances this field with improved performance and efficiency~\cite{KevisKokitsiManinis2017DeepEC,ZhuwenLi2018InteractiveIS,ZhengLin2020InteractiveIS,JangWonDong2019InteractiveIS,KonstantinSofiiuk2020fBRSRB,DavidAcuna2018EfficientIA,XiChen2022FocalClickTP,QinLiu2022PseudoClickII,QinLiu2022SimpleClickII,huang2023interformer,xu2024structured,zhang2024leveraging}. The introduction of SAM~\cite{kirillov2023segany} marks a significant advancement in inference efficiency through feature reuse, and inspires numerous downstream applications~\cite{ma2023segment,mazurowski2023segment,lai2023lisa}. This evolution leads to the emergence of high-precision interactive segmentation~\cite{liu2024rethinking,huang2024hrsam,wang2024order}, which aims for superior accuracy on precisely annotated datasets~\cite{FedericoPerazzi2016ABD,sam_hq}.

\noindent\textbf{Vector Quantized Representation Learning.} The Vector Quantization Variational AutoEncoder (VQ-VAE)~\cite{van2017neural} pioneered discrete representation learning through learnable codebook quantization. Later works address codebook collapse via perceptual losses~\cite{esser2021taming}, $\ell_2$ normalization~\cite{yu2021vector}, implicit codebooks~\cite{mentzer2023finite}, factorized quantization~\cite{el2022image}, and dynamic updates~\cite{zheng2023online}. Most notably, BSQ~\cite{zhao2024image} introduces a parameter-free approach by projecting and binarizing latent vectors on a hypersphere, enabling efficient tokenization for vision transformers.

\noindent\textbf{Efficient Attention.}
Attention~\cite{AshishVaswani2017AttentionIA} has revolutionized computer vision~\cite{AlexeyDosovitskiy2020AnII,khan2022transformers,fang2023eva,wang2023internimage}. However, its quadratic computational complexity motivates continuous exploration of efficient solutions. Previous works improve efficiency through local windows~\cite{liu2021swin}, sparse patterns~\cite{Guo_Qiu_Liu_Shao_Xue_Zhang_2019}, and hierarchical structures~\cite{YanghaoLiExploringPV,han2023flatten}. At the implementation level, Flash Attention~\cite{dao2022flashattention,dao2023flashattention2} and EMA~\cite{DBLP:journals/corr/abs-2112-05682} reduce memory footprint through optimized computation scheduling. Most recently, Transformer-VQ~\cite{lingle2023transformer} achieves linear time complexity by representing keys through vector quantization, where attention only needs to be computed between queries and a fixed-size codebook rather than the full key sequence.

\section{Method}

\subsection{Background}
\label{subsec:background}

\red\noindent\textbf{Interactive segmentation.}
Interactive segmentation (IS) performs foreground-background segmentation using user prompts. Given an input image $\mathbf{I} \in \mathbb{R}^{3\times H \times W}$ and $n$ user clicks $C_n = \{(y_i, x_i, z_i)\}_{i=1}^n$, where $(y_i, x_i) \in [0, H - 1] \times [0, W - 1]$ denotes click coordinates and $z_i \in \{0, 1\}$ indicates positive (foreground) or negative (background) clicks, the IS model outputs a binary mask $\mathbf{M} \in \{0, 1\}^{H \times W}$. While bounding boxes and scribbles are alternative prompts, they can be converted to clicks. Click-based IS has become mainstream due to standardized evaluation and high segmentation accuracy (\eg IoU $>95\%$).

\noindent\textbf{Efficient IS pipeline.}
Traditional IS models convert clicks into click maps, which are two-channel binary masks matching the image size. Each channel stores positive or negative clicks by representing each click as a circular disk, where pixels within radius $R$ (\eg $R=5$) from the click center are set to 1 and all other pixels are set to 0. These click maps combine with the input image to form a 5-channel input for vision model inference. However, this process creates computational redundancy through repeated image processing. SAM~\cite{kirillov2023segany} and InterFormer~\cite{huang2023interformer} improve efficiency by decoupling the IS pipeline into two stages: image preprocessing before interaction and lightweight decoding during interaction. In the preprocessing stage, a ViT encoder converts the input image into $d$-dimensional image tokens $\mathbf{F} = \text{Encoder}(\mathbf{I}) \in \mathbb{R}^{d \times h \times w}$, where $h = H/16, w = W/16$. During interaction, at step $k$, a lightweight decoder processes these tokens along with current clicks $\mathbf{C}_k$ and the previous segmentation mask $\mathbf{M}_{k-1}$ to generate the refined mask $\mathbf{M}_k = \text{Decoder}(\mathbf{F}, \mathbf{C}_k, \mathbf{M}_{k-1})$.

\noindent\textbf{Different decoder designs.}
SAM and InterFormer differ in decoder design: SAM encodes $C_n$ into sparse prompt tokens for efficiency, while InterFormer uses click maps as dense prompt tokens for accuracy. This paper builds upon InterFormer's architecture and click maps to propose Inter2Former, optimizing dense prompt token computation to achieve SAM-level efficiency.

\subsection{Overview of Inter2Former}
\label{subsec:i2foverview}

\noindent\textbf{Encoder.}
Inter2Former adopts HRSAM's encoder~\cite{huang2024hrsam}, utilizing the Flash Swin and multi-scale fusion to generate image tokens $\mathbf{F} \in \mathbb{R}^{d \times h \times w}$. Considering that HRSAM's Selective State Space (SSM) operator~\cite{gu2023mamba} cannot be computed on CPU, we simplify the HRSAM encoder by removing SSM while maintaining its visual extrapolation capability~\cite{huang2024hrsam} for high-resolution images.

\noindent\textbf{Decoder.}
As illustrated in Figure~\ref{fig:overview}, the Inter2Former decoder comprises three main components. First, following InterFormer's prompt embedding scheme~\cite{huang2023interformer}, user clicks are encoded into click maps and combined with the previous mask to generate prompt tokens $\mathbf{P} \in \mathbb{R}^{d\times h \times w}$, which are dimensionally aligned with image tokens $\mathbf{F}$ via our efficient DPE that focuses on local regions. Second, adopting the standard transformer architecture~\cite{AshishVaswani2017AttentionIA}, we process the prompt-fused tokens $\mathbf{F}_P = \mathbf{F} + \mathbf{P}$ through alternating layers of our DHA (Attention) and HMoE (FFN) modules to obtain refined tokens $\mathbf{F}_R \in \mathbb{R}^{d\times h \times w}$. Finally, we leverage DLU to upsample $\mathbf{F}_R$ into the final segmentation mask.

\subsection{Dynamic Prompt Embedding}
\label{subsec:dpe}

Following InterFormer~\cite{huang2023interformer}, we maintain a reference mask $\mathbf{M}_{\text{ref}} \in \{0,1,2,3,4\}^{H \times W}$ to encode user clicks and previous predictions. In $\mathbf{M}_{\text{ref}}$, values $0$ and $4$ mark definite background and foreground regions within radius-5 circles of clicks. Values $1$ and $3$ indicate possible background and foreground from predictions. Value $2$ represents uncertain regions with unstable predictions. During interaction, we update $\mathbf{M}_{\text{ref}}$ using InterFormer's rules based on new clicks and predictions.

To reduce computational cost, we propose Dynamic Prompt Embedding (DPE) for efficient $\mathbf{M}_{\text{ref}}$ encoding. We first detect a bounding box $\mathcal{B} = [x_1:x_2, y_1:y_2]$ containing all click regions and foreground predictions with padding. For the local region of $\mathcal{B}$, we apply learnable embedding:
\begingroup
\setlength{\abovedisplayskip}{0.5em}
\setlength{\belowdisplayskip}{0.5em}
\begin{equation}
    \mathbf{E}_{\text{init}} = \text{Embed}(\mathbf{M}_{\text{ref}}[y_1:y_2, x_1:x_2]) \in \mathbb{R}^{5 \times H_{\mathcal{B}} \times W_{\mathcal{B}}},
\end{equation}
\endgroup
where $H_{\mathcal{B}} = y_2 - y_1$, $W_{\mathcal{B}} = x_2 - x_1$, and $\text{Embed}(\cdot)$ maps each value to a learnable 5-dimensional vector. The features are downsampled through four stride-2 convolutions:
\begingroup
\setlength{\abovedisplayskip}{0.5em}
\setlength{\belowdisplayskip}{0.5em}
\begin{equation}
    \mathbf{F}_{\mathcal{B}} = \text{Conv}_4(\mathbf{E}_{\text{init}}) \in \mathbb{R}^{d \times h_{\mathcal{B}} \times w_{\mathcal{B}}},
\end{equation}
\endgroup
where $h_{\mathcal{B}} = H_{\mathcal{B}}/16$ and $w_{\mathcal{B}} = W_{\mathcal{B}}/16$. The final prompt embedding combines local features with a learnable background embedding $\mathbf{e}_{\text{bg}}$:
\begingroup
\setlength{\abovedisplayskip}{0.5em}
\setlength{\belowdisplayskip}{0.5em}
\begin{equation}
    \mathbf{P} = \begin{bmatrix} 
    \mathbf{e}_{\text{bg}} & \cdots & \mathbf{e}_{\text{bg}} \\
    \vdots & \mathbf{F}_{\mathcal{B}} & \vdots \\
    \mathbf{e}_{\text{bg}} & \cdots & \mathbf{e}_{\text{bg}}
    \end{bmatrix} \in \mathbb{R}^{d \times h \times w}.
\end{equation}
\endgroup
This design reduces computation by processing only regions of interest while maintaining global context through background embedding.

\subsection{Dynamic Hybrid Attention}
\label{subsec:dha}

\noindent\textbf{Computation Allocation.}
To efficiently allocate computation at interaction step $k$, we identify object boundaries from the previous mask $\mathbf{M}_{k-1}$. We first apply a $7\times7$ average convolution (uniform weights $1/49$) to estimate local variance and detect non-zero variance regions. The resulting edge map is then downsampled via max pooling to match the dimensions of $\mathbf{F}_P$:
\begingroup
\setlength{\abovedisplayskip}{0.5em}
\setlength{\belowdisplayskip}{0.5em}
\begin{equation}
{\fontsize{9.25}{11}\selectfont
    \begin{split}
    \mathbf{E}_{k-1} =
    \text{Pool}\left(\mathbbm{1}\{\text{Conv}(\mathbf{M}_{k-1}^2) - \text{Conv}(\mathbf{M}_{k-1})^2 > 0\}\right).
    \end{split}
}
    \label{eq:edge}
\end{equation}
\endgroup
Guided by the edge map $\mathbf{E}_{k-1}$, we allocate computational resources as follows. The prompt-fused tokens $\mathbf{F}_P$ are reshaped into sequence form $\mathbf{F}_{\text{flat}} = \text{Reshape}(\mathbf{F}_P) \in \mathbb{R}^{L \times d}$, where $L = h \times w$. Linear projections are then applied to generate query, key and value matrices: $(\mathbf{Q}, \mathbf{K}, \mathbf{V}) = (\mathbf{F}_{\text{flat}}\mathbf{W}_q, \mathbf{F}_{\text{flat}}\mathbf{W}_k, \mathbf{F}_{\text{flat}}\mathbf{W}_v) \in \mathbb{R}^{L \times C}$. Based on $\mathbf{E}_{k-1}$, we partition queries into two disjoint sets:
\begingroup
\setlength{\abovedisplayskip}{0.5em}
\setlength{\belowdisplayskip}{0.5em}
\begin{equation}
    \begin{split}
        \mathbf{Q}_{\text{FA}} &= \{\mathbf{q}_i | \mathbf{E}_{k-1}(i) = 1\} \\
        \mathbf{Q}_{\text{BSQ}} &= \{\mathbf{q}_i | \mathbf{E}_{k-1}(i) = 0\}
    \end{split}
\end{equation}
\endgroup
where $\mathbf{Q}_{\text{FA}}$ corresponds to edge regions (minority) and $\mathbf{Q}_{\text{BSQ}}$ to non-edge regions (majority). Both sets share the same key-value pairs $(\mathbf{K}, \mathbf{V})$ but employ different attention computations: $\mathbf{Q}_{\text{FA}}$ uses full attention with $O(N^2)$ complexity, while $\mathbf{Q}_{\text{BSQ}}$ adopts BSQA with $O(N)$ complexity, thus balancing effectiveness and efficiency.

\noindent\textbf{Full Attention.}
For queries in $\mathbf{Q}_{\text{FA}}$, we apply the standard full attention computation. Given the shared key-value pairs $(\mathbf{K}, \mathbf{V}) \in \mathbb{R}^{N \times C}$, where $C$ is the head dimension (we present the single-head case for clarity, while our implementation adopts standard multi-head attention as in transformers), the attention output is computed as:
\begingroup
\setlength{\abovedisplayskip}{0.5em}
\setlength{\belowdisplayskip}{0.5em}
\begin{equation}
    \begin{split}
        \text{FA}(\mathbf{Q}_{\text{FA}}) = \text{Softmax}\left(\frac{\mathbf{Q}_{\text{FA}}\mathbf{K}^\top}{\sqrt{C}}\right)\mathbf{V}.
    \end{split}
    \label{eq:attention}
\end{equation}
\endgroup
This FA computation has quadratic complexity $O(N^2)$ with respect to sequence length. For the majority of queries in $\mathbf{Q}_{\text{BSQ}}$, we introduce our proposed BSQA that achieves linear complexity while maintaining attention effectiveness.

\subsection{Binary Spherical Quantization Attention}
\label{subsec:bsqa}
Inspired by VQ-Transformer~\cite{lingle2023transformer}, we propose BSQA that enhances VQ Attention with a more effective Binary Spherical Quantization (BSQ) scheme while maintaining its linear-time complexity. Below we elaborate on VQ Attention, BSQ and our proposed BSQA respectively.

\noindent\textbf{Vector Quantized Attention.}
The insight of VQ Attention is applying vector quantization to the $N$ key vectors $\mathbf{K} \in \mathbb{R}^{N \times C}$ through a learned codebook $\mathbf{C} \in \mathbb{R}^{S \times C}$ of size $S$. Each key vector is mapped to its nearest codebook vector:
\begingroup
\setlength{\abovedisplayskip}{0.5em}
\setlength{\belowdisplayskip}{0.5em}
\begin{equation}
    \begin{split}
        \hat{\mathbf{K}} = \text{VQ}(\mathbf{K}, \mathbf{C}) = \text{arg}\min_{\mathbf{c}_i \in \mathbf{C}} \|\mathbf{k} - \mathbf{c}_i\|_2^2.
    \end{split}
    \label{eq:vq_nearest}
\end{equation}
\endgroup
This quantization process can be expressed through a binary assignment matrix $\boldsymbol{\Delta} \in \{0,1\}^{N \times S}$, where $\hat{\mathbf{K}} = \mathbf{\Delta}\mathbf{C}$. The attention computation (without softmax's normalization) can then be reformulated as:
\begingroup
\setlength{\abovedisplayskip}{0.5em}
\setlength{\belowdisplayskip}{0.5em}
\begin{equation}
    \begin{split}
        \text{exp}(\mathbf{Q}\hat{\mathbf{K}}^\top)\mathbf{V} = \text{exp}(\mathbf{Q}\mathbf{C}^\top)(\boldsymbol{\Delta}^\top\mathbf{V}).
    \end{split}
    \label{eq:vq_attention}
\end{equation}
\endgroup
The binary nature of $\mathbf{\Delta}$ enables this equivalent factorization, which reduces the computational complexity from quadratic $O(N^2)$ to linear $O(NS) = O(N)$ with fixed $S$. 
The complete attention with softmax normalization can be computed by applying the same factorization twice: once with the $\mathbf{V}$, and once with an all-ones vector $\mathbf{1}_{N\times 1}$:
\begingroup
\setlength{\abovedisplayskip}{0.5em}
\setlength{\belowdisplayskip}{0.5em}
\begin{equation}
    \text{Attention}(\mathbf{Q}, \mathbf{K}, \mathbf{V}) = \frac{\text{exp}(\mathbf{Q}\mathbf{C}^\top)(\boldsymbol{\Delta}^\top\mathbf{V})}{\text{exp}(\mathbf{Q}\mathbf{C}^\top)(\boldsymbol{\Delta}^\top\mathbf{1}_{N\times 1})}.
    \label{eq:vq_attention_full}
\end{equation}
\endgroup
However, nearest-neighbor quantization $\text{VQ}(\cdot, \cdot)$ tends to underutilize the codebook by activating only a small subset of vectors. More critically, when applying straight-through estimation (STE), the gradient approximation relies on copying gradients from quantized $\hat{\mathbf{K}}$ to $\mathbf{K}$:
\begingroup
\setlength{\abovedisplayskip}{0.5em}
\setlength{\belowdisplayskip}{0.5em}
\begin{equation}
    \frac{\partial \mathcal{L}}{\partial \mathbf{K}} \approx \frac{\partial \mathcal{L}}{\partial \hat{\mathbf{K}}}.
    \label{eq:ste}
\end{equation}
\endgroup
Although minimizing $\|\hat{\mathbf{K}} - \mathbf{K}\|_2^2$ is introduced as a commitment loss to bound this approximation, the quantization error remains uncontrollable during training. To address these fundamental limitations, we adopt BSQ instead.

\noindent\textbf{Binary Spherical Quantization.} 
BSQ~\cite{zhao2024image} addresses the limitations of VQ by projecting features onto a unit hypersphere before quantization. Likewise, given $N$ key vectors $\mathbf{K} \in \mathbb{R}^{N \times C}$, BSQ first maps it to a lower-dimensional space through a learnable transformation:
\begingroup
\setlength{\abovedisplayskip}{0.5em}
\setlength{\belowdisplayskip}{0.5em}
\begin{equation}
    \mathbf{B} = \mathbf{K}\mathbf{W}_{\text{BSQ}} \in \mathbb{R}^{N \times S} \text{ where } S \ll C.
    \label{eq:bsq_project}
\end{equation}
\endgroup
The projected vector is then normalized onto a unit sphere:
\begingroup
\setlength{\abovedisplayskip}{0.5em}
\setlength{\belowdisplayskip}{0.5em}
\begin{equation}
    \mathbf{U} = \mathbf{B}/\|\mathbf{B}\|_2.
    \label{eq:bsq_norm}
\end{equation}
\endgroup
Unlike VQ's nearest-neighbor assignment, BSQ applies element-wise binary quantization:
\begingroup
\setlength{\abovedisplayskip}{0.5em}
\setlength{\belowdisplayskip}{0.5em}
\begin{equation}
    \hat{\mathbf{U}} = \text{sign}(\mathbf{U})/\sqrt{S},
    \label{eq:bsq_quantize}
\end{equation}
\endgroup
where the scaling factor ensures the quantized vector maintains unit norm. The quantized features are then projected back to the original dimension through another learnable transformation, which will be discussed in the next section.
Since BSQ's quantization error $\|\mathbf{U} - \hat{\mathbf{U}}\|$ is theoretically bounded~\cite{zhao2024image} and empirically converges close to zero during our experiments, it enables accurate gradient estimation through STE, addressing VQ's gradient approximation challenge. Next, we will introduce BSQA, which further addresses the inefficient codebook utilization issue in VQ Attention.

\noindent\textbf{BSQ Attention.}
To leverage $\hat{\mathbf{U}}$ in attention computation similar to VQ Attention's codebook vectors, we first transform $\hat{\mathbf{U}}$ through a simple linear mapping: $\mathbf{I} = \hat{\mathbf{U}} \times \sqrt{S} / 2 + 1/2 \in \{0, 1\}^{N\times S}$, which encodes each key as an $S$-bit binary representation. We then construct the final projected vectors by combining learnable base vectors $\mathbf{C}_{\text{base}}^0, \mathbf{C}_{\text{base}}^1 \in \mathbb{R}^{S \times C}$. Specifically, for each bit position $j$, we select either $\mathbf{C}_{\text{base}}^0$ or $\mathbf{C}_{\text{base}}^1$ based on the binary value, and aggregate these selections across all bit positions. This aggregation process can be efficiently formulated as
\begingroup
\setlength{\abovedisplayskip}{0.5em}
\setlength{\belowdisplayskip}{0.5em}
\begin{equation}
    \hat{\mathbf{K}} = [\mathbf{I} \quad \mathbf{1} - \mathbf{I}] 
    \begin{bmatrix}
        \mathbf{C}_{\text{base}}^1 \\
        \mathbf{C}_{\text{base}}^0
    \end{bmatrix}
    \in \mathbb{R}^{N \times C}.
    \label{eq:bsqa}
\end{equation}
\endgroup
The $S$-bit binary representation allows $2^S$ different combinations $\{0, 1, \ldots, 2^S - 1\}$. Through our binary selection mechanism with $\mathbf{C}_{\text{base}}^0$ and $\mathbf{C}_{\text{base}}^1$, we naturally construct a codebook of $2^S$ vectors. Following the formulation similar to Eq.~\ref{eq:bsqa}, these vectors constitute the codebook required for VQ Attention, enabling computations analogous to Eq.~\ref{eq:vq_attention}.

\subsection{Hybrid Mixture of Experts}
\label{subsec:hmoe}

HMoE builds upon DeepSeek V3's MoE design~\cite{liu2024deepseek} including its auxiliary-loss-free expert balancing strategy. Like DHA, HMoE adopts a hybrid strategy where the computation path for each token is determined by the edge map $\mathbf{E}_{k-1}$. With $M$ routed experts $\{\text{FFN}_i\}_{i=0}^{M-1}$ and one shared expert $\text{FFN}_M$, for input tokens $\mathbf{X} \in \mathbb{R}^{N \times d}$, non-edge tokens where $\mathbf{E}_{k-1}(t)=0$ only utilize the shared expert:
\begingroup
\setlength{\abovedisplayskip}{0.5em}
\setlength{\belowdisplayskip}{0.5em}
\begin{equation}
    \mathbf{y}_t = \text{FFN}_M(\mathbf{x}_t).
\end{equation}
\endgroup
For edge tokens where $\mathbf{E}_{k-1}(t)=1$, we first compute token-expert affinity scores:
\begingroup
\setlength{\abovedisplayskip}{0.5em}
\setlength{\belowdisplayskip}{0.5em}
\begin{equation}
    s_{i,t} = \text{Sigmoid}(\mathbf{x}_t^\top\mathbf{e}_i), 0 \le i \le M,
\end{equation}
\endgroup
where $\mathbf{e}_i$ represents the learnable centroid vector of the $i$-th expert. The token then selects the expert with highest affinity among $M$ routed experts:
\begingroup
\setlength{\abovedisplayskip}{0.5em}
\setlength{\belowdisplayskip}{0.5em}
\begin{equation}
    a_t = \text{arg}\max_{0 \le i \le M-1} s_{i,t},
\end{equation}
\endgroup
and combines its output with the shared expert:
\begingroup
\setlength{\abovedisplayskip}{0.5em}
\setlength{\belowdisplayskip}{0.5em}
\begin{equation}
    \small
    \mathbf{y}_t = \frac{\exp(s_{M,t})\text{FFN}_M(\mathbf{x}_t) + \exp(s_{a_t,t})\text{FFN}_{a_t}(\mathbf{x}_t)}{\exp(s_{M,t}) + \exp(s_{a_t,t})}.
\end{equation}
\endgroup

\noindent\textbf{Efficient Parallel Processing.} 
To enable efficient matrix operations for expert computation, we propose a token rearrangement strategy specifically for edge tokens. Using their selected expert indices $\{a_t | \mathbf{E}_{k-1}(t)=1\}$ where $a_t \in \{0,1,\ldots,M-1\}$, we first sort these tokens to obtain contiguous groups:
\begingroup
\setlength{\abovedisplayskip}{0.5em}
\setlength{\belowdisplayskip}{0.5em}
\begin{equation}
    \pi = \text{argsort}(\{a_t | \mathbf{E}_{k-1}(t)=1\}),
\end{equation}
\endgroup
where $\pi$ is the permutation that sorts edge tokens by expert ID. The sorted tokens form $M$ groups:
\begingroup
\setlength{\abovedisplayskip}{0.5em}
\setlength{\belowdisplayskip}{0.5em}
\begin{equation}
    \mathbf{X}_i = \{\mathbf{x}_{\pi[j]} | a_{\pi[j]} = i\} \in \mathbb{R}^{N_i \times d},
\end{equation}
\endgroup
where $N_i$ is the number of edge tokens assigned to routed expert $i$. By rearranging tokens assigned to the same expert into contiguous memory blocks, we can decompose each expert's FFN computation into matrix multiplications (as Linear layers are essentially matrix multiplications) with activation functions. Through C++ extension with low-level matrix operation optimization and batch parallelism, these blocked matrix operations for multiple experts are efficiently executed simultaneously. The final outputs are recovered by inverse permutation $\pi^{-1}$. This multi-expert batch processing scheme achieves significant speedup over sequential computation.

\subsection{Dynamic Local Upsampling}
\label{subsec:dlu}

\noindent\textbf{Local Refinement.}
We apply DLU to generate the segmentation mask from tokens $\mathbf{F}_R \in \mathbb{R}^{d\times h \times w}$ refined by DHA and HMoE modules. Our DLU consists of a localization branch and a refinement branch for efficient computation allocation. The localization branch first generates a coarse mask without upsampling to identify the object region:
\begingroup
\setlength{\abovedisplayskip}{0.5em}
\setlength{\belowdisplayskip}{0.5em}
\begin{equation}
    \mathbf{M}_{\text{low-res}} = \text{MLP}(\mathbf{F}_R; d \rightarrow 1) \in \mathbb{R}^{1 \times h \times w},
    \label{eq:lowres}
\end{equation}
\endgroup
from which we extract an expanded bounding box $\mathcal{B} = [x_1:x_2, y_1:y_2]$ around detected objects. The refinement branch then focuses on precise boundary delineation within the detected region. We first extract tokens within the expanded bounding box $\mathcal{B}$, where $h_{\mathcal{B}} = y_2 - y_1$ and $w_{\mathcal{B}} = x_2 - x_1$ define the local region dimensions:
\begingroup
\setlength{\abovedisplayskip}{0.5em}
\setlength{\belowdisplayskip}{0.5em}
\begin{equation}
    \mathbf{F}_{\mathcal{B}} = \mathbf{F}_R[:, y_1:y_2, x_1:x_2] \in \mathbb{R}^{d \times h_{\mathcal{B}} \times w_{\mathcal{B}}}.
    \label{eq:highres}
\end{equation}
\endgroup
Based on the extracted features $\mathbf{F}_{\mathcal{B}}$, the refinement branch performs precise boundary delineation through edge-guided upsampling, where edge features serve as complementary boundary information.

\noindent\textbf{Edge-guided Upsampling.}
We first extract multi-scale edge features through a lightweight CannyNet prior to interaction:
\begingroup
\setlength{\abovedisplayskip}{0.5em}
\setlength{\belowdisplayskip}{0.5em}
\begin{equation}
    \begin{split}
        \mathbf{E} &= \text{Canny}(\mathbf{I}), \\
        \{\mathbf{F}^e_i\}_{i=1}^4 &= \text{ResNet}(\mathbf{E}),
    \end{split}
\end{equation}
\endgroup
where $\mathbf{E}$ denotes the binary edge map and $\{\mathbf{F}^e_i\}_{i=1}^4$ represents edge features at increasing scales. These features are fused with $\mathbf{F}_{\mathcal{B}}$ during upsampling to generate the final mask:
\begingroup
\setlength{\abovedisplayskip}{0.5em}
\setlength{\belowdisplayskip}{0.5em}
\begin{equation}
    \mathbf{M} = \begin{bmatrix} 
    0 & \cdots & 0 \\
    \vdots & \text{Up}(\mathbf{F}_{\mathcal{B}}, \{\mathbf{F}^e_i\}_{i=1}^4) & \vdots \\
    0 & \cdots & 0
    \end{bmatrix} \in \mathbb{R}^{H \times W},
    \label{eq:ups}
\end{equation}
\endgroup
where $\text{Up}(\cdot)$ consists of four deconvolution layers ($\times2$) that progressively expand resolution while reducing channels to 1. At each upsampling stage, the corresponding $\mathbf{F}^e_i$ is fused with the upsampled feature map via addition, followed by a convolutional layer for feature refinement.

\subsection{Training Strategy}
\label{subsec:training}

\noindent\textbf{BSQA Training.}
During training, we replace keys in attention computation using Eq.~\ref{eq:bsqa} while performing complete FA computation, only with the quantized key vectors. This training strategy encourages the quantized attention to closely approximate the behavior of standard attention. At inference time, we switch to the efficient computation scheme in Eq.~\ref{eq:vq_attention} to achieve linear-time complexity.

\noindent\textbf{DLU Training.} 
The DLU module generates two outputs during training: a low-resolution mask $\mathbf{M}_{\text{low-res}}$ from Eq.~\ref{eq:lowres} and a high-resolution mask $\mathbf{M}$ by applying $\text{Up}(\cdot)$ from Eq.~\ref{eq:highres} to the full token set. These outputs are supervised by downsampled and original GT masks respectively, using the NFL loss~\cite{KonstantinSofiiuk2021RevivingIT} common in interactive segmentation. At inference, we employ DLU for efficient mask generation.

\begin{table*}[!t]
  \centering
  \resizebox{\textwidth}{!}{
  \begin{tabu}[c]{l c c c c c c c c}
    \toprule
    \multirow{2}{*}{\textbf{Model}} & 
    \multirow{2}{*}{\textbf{Training Data}} &
    \textbf{CPU Time (ms) $\downarrow$} &
    \multicolumn{3}{c}{\textbf{HQSeg44K} {\small$_{\textbf{Max H/W} > 4000}$}} & 
    \multicolumn{3}{c}{\textbf{DAVIS} {\small$_{\textbf{Max H/W} < 1000}$}} \\
    \cmidrule(lr){4-6} \cmidrule(lr){7-9}
    & & \textbf{$^\ddagger$20-SPC / Online} & 
    5-mIoU $\uparrow$ & NoC90 $\downarrow$ & NoC95 $\downarrow$ &
    5-mIoU $\uparrow$ & NoC90 $\downarrow$ & NoC95 $\downarrow$ \\
    \midrule
    RITM-HRNet32 $_{400}$~\cite{KonstantinSofiiuk2021RevivingIT} & COCO+LVIS & 277 / 277 &
    77.72 & 10.01 & 14.58 & 89.75 & 5.34 & 11.45 \\
    FocalClick-SegF-B3-S2 $_{256}$~\cite{XiChen2022FocalClickTP} & COCO+LVIS & 97 / 97 &
    84.63 & 8.12 & 12.63 & 90.82 & 5.17 & 11.42 \\
    FocalClick-SegF-B3-S2 $_{384}$~\cite{XiChen2022FocalClickTP} & COCO+LVIS & 175 / 175 &
    85.45 & 7.03  & 10.74 & 91.22 & 4.90 & 10.40 \\    
    SimpleClick-ViT-B $_{448}$~\cite{QinLiu2022SimpleClickII} & COCO+LVIS & 212 / 212 &
    85.11 & 7.47  & 12.39 & 90.73 & 5.06 & 10.37 \\    
    InterFormer-ViT-B $_{1024}$~\cite{huang2023interformer} & COCO+LVIS & 1020 / 188 &
    82.62 & 7.17  & 10.77 & 87.79 & 5.45 & 11.88 \\
    SegNext (SA$\times$1) ViT-B $_{1024}$~\cite{liu2024rethinking} & COCO+LVIS & 910 / 803 &
    85.41 & 7.47 & 11.94 & 90.13 & 5.46 & 13.31 \\    
    SegNext (SA$\times$2) ViT-B $_{1024}$~\cite{liu2024rethinking} & COCO+LVIS & 1519 / 1400 &
    85.71 & 7.18 & 11.52 & 89.85 & 5.34 & 12.80 \\
    SegNext (SA$\times$2) ViT-B $_{1024}$~\cite{liu2024rethinking} & COCO+LVIS+HQ & 1519 / 1400 &
    91.75 & 5.32 & 9.42 & 91.87 & 4.43 & 10.73 \\
    HRSAM$^{++}$-ViT-B $_{1024}$~\cite{huang2024hrsam} & COCO+LVIS+HQ 
    & 65 / 40 & 90.32 & 6.27 & 10.14 & 90.40 & 5.71 & 12.72 \\
    HRSAM$^{++}$-ViT-B $_{2048}$~\cite{huang2024hrsam} & COCO+LVIS+HQ 
    & 273 / 105 & 91.50 & 5.41 & 9.08 & 90.79 & 5.52 & 10.84 \\
    \midrule
    SAM-ViT-B $_{1024}$~\cite{kirillov2023segany} & SA-1B & 142 / 40 &
    86.16 & 7.46 & 12.42 & 90.95 & 5.14 & 10.74 \\
    MobileSAM-ViT-T $_{1024}$~\cite{zhang2023faster} & SA-1B & 35 / 24 &
    81.98 & 8.70 & 13.83 & 89.18 & 5.83 & 12.74 \\    
    EfficientSAM-ViT-T $_{1024}$~\cite{xiong2023efficientsam} & ImageNet+SA-1B & 88 / 27 &
    77.90 & 10.11 & 14.60 & 85.26 & 7.37 & 14.28 \\ 
    EfficientSAM-ViT-T $_{2048}$~\cite{xiong2023efficientsam} & ImageNet+SA-1B & 706 / 77 &
    74.20 & 9.47 & 13.13 & 84.10 & 8.00 & 14.37 \\ 
    EfficientSAM-ViT-S $_{1024}$~\cite{xiong2023efficientsam} & ImageNet+SA-1B & 127 / 23 &
    79.01 & 8.84 & 13.18 & 87.55 & 6.37 & 12.26 \\ 
    EfficientSAM-ViT-S $_{2048}$~\cite{xiong2023efficientsam} & ImageNet+SA-1B & 1658 / 69 &
    74.91 & 8.27 & 11.97 & 85.17 & 6.86 & 12.49 \\ 
    HQ-SAM-ViT-B $_{1024}$~\cite{ma2023segment} & SA-1B+HQ & 167 / 54 &
    89.85 & 6.49 & 10.79 & 91.77 & 5.26 & 10.00 \\
    HRSAM$^{++}$-ViT-B $_{1024}$~\cite{huang2024hrsam} & $^\dagger$SA-1B+HQ 
    & 65 / 40 & 89.37 & 6.56 & 10.61 & 86.66 & 7.29 & 14.32 \\
    HRSAM$^{++}$-ViT-B $_{2048}$~\cite{huang2024hrsam} & $^\dagger$SA-1B+HQ 
    & 273 / 105 & 90.94 & 5.86 & 9.22 & 88.46 & 6.61 & 12.39 \\
    \midrule
    Inter2Former-ViT-B$_{1024}$ (Ours) & COCO+LVIS+HQ 
    & 75 / 50 & 91.48 & 5.36 & 9.29 & 90.82 & 4.90 & 11.33 \\
    Inter2Former-ViT-B$_{2048}$ (Ours) & COCO+LVIS+HQ 
    & 300 / 131 & 92.28 & 4.58 & 7.79 & 91.30 & 4.33 & 8.45 \\
    Inter2Former-ViT-B$_{1024}$ (Ours) & $^\dagger$SA-1B+HQ 
    & 75 / 50 & 91.32 & 5.46 & 9.34 & 90.36 & 4.96 & 12.72 \\
    Inter2Former-ViT-B$_{2048}$ (Ours) & $^\dagger$SA-1B+HQ 
    & 300 / 131 & \textbf{92.68} & \textbf{4.24} & \textbf{7.39} & \textbf{92.00} & \textbf{4.29} & \textbf{7.82} \\
    \bottomrule
  \end{tabu}}
  \vspace{-0.5em}
  \caption{
  Performance evaluation over high-precision IS tasks. We compare our Inter2Former with SOTA methods on HQSeg44K and DAVIS datasets under two training protocols: (1) conventional COCO+LVIS training with HQ fine-tuning, and (2) distillation from SAM followed by HQ training ($\dagger$ indicates models trained via this protocol). For efficiency, we report CPU inference time ($\ddagger$ where 20-SPC evaluates the average time for standard 20 clicks including preprocessing, \eg SAM's image encoding, while Online measures per-click latency during interaction excluding preprocessing). While Inter2Former shows marginally higher latency compared to sparse-token-based HRSAM, it demonstrates superior segmentation quality and achieves SOTA performance across 5-mIoU and NoC@90/95.
}
  \label{tab:quantitative_comparison}
\end{table*}

\section{Experiments}
Section~\ref{subsec:implementation} details our implementation. Section~\ref{subsec:experimentalsetting} describes datasets and training protocols. Section~\ref{subsec:mainresult} presents the main quantitative comparison. Section~\ref{subsec:efficiency} analyzes computational efficiency of our key components. Additionally, we provide the \textbf{code} in the supplementary materials.

\subsection{Implementation Details}
\label{subsec:implementation}

Our Inter2Former adopts the encoder from HRSAM++ \cite{huang2024hrsam}, specifically a ViT-Base with 12 layers, 768 embedding dimensions, and 12 attention heads, initialized with MAE pre-training~\cite{he2017mask}. For CPU compatibility, we remove the Cycle-scan module (with SSM~\cite{gu2023mamba}) from the original HRSAM++ while retaining its anchor map (fixed size of 512). For fair comparison, we also re-train HRSAM++ without SSM under the same settings.

The decoder of Inter2Former consists of two alternating layers of DHA and HMoE modules. While it processes 256-dimensional input tokens, the attention computation is performed with reduced 32-dimensional Q/K/V vectors for efficiency. We enhance FA with Rotary Position Embedding (RoPE)~\cite{su2024roformer} and implement BSQA using an $8$-bit codebook (16 base vectors). The DPE module transforms the original-scale input through four convolutional layers into $256$-dimensional prompt tokens at $1/16$ resolution, while DLU reverses this process through four convolutions to generate single-channel full-resolution masks. The proposed CannetNet adopts a straightforward design with four stages of dual ResNet blocks~\cite{he2016deep}, processing features of dimensions 4, 16, 64 and 256 respectively, with each stage halving the spatial resolution.

\subsection{Experimental Setting}
\label{subsec:experimentalsetting}

\noindent\textbf{Datasets.} 
Following recent high-precision interactive segmentation benchmarks~\cite{liu2024rethinking,huang2024hrsam}, we train our model on COCO~\cite{TsungYiLin2014MicrosoftCC}, LVIS~\cite{AgrimGupta2019LVISAD} and HQSeg-44K~\cite{sam_hq} datasets. We evaluate on HQSeg-44K validation set and DAVIS~\cite{FedericoPerazzi2016ABD}, both of which contain high-quality annotated masks for rigorous evaluation over high-precision IS tasks.

\noindent\textbf{Training.} Following previous high-precision IS works, we adopt two training protocols for comprehensive evaluation. The first protocol follows~\cite{liu2024rethinking}, where we train on COCO and LVIS for 160K iterations, followed by 40K iterations of fine-tuning on HQSeg44K. 
The second protocol involves distillation from SAM-ViT-Huge~\cite{sam_hq} where we first train the encoder for 160K iterations on unlabeled images from COCO and LVIS using MSE loss to align with SAM features, then randomly initialize the Inter2Former decoder and train the complete model for 80K iterations on HQSeg44K. Both protocols utilize click simulations from prior works~\cite{KonstantinSofiiuk2021RevivingIT,huang2023interformer,QinLiu2022SimpleClickII} and NFL~\cite{KonstantinSofiiuk2021RevivingIT} to guide the model in producing GT-aligned segmentation masks from clicks.


\noindent\textbf{Evaluation.} Following standard evaluation protocols~\cite{KonstantinSofiiuk2021RevivingIT,XiChen2022FocalClickTP,QinLiu2022SimpleClickII,huang2023interformer,liu2024rethinking,kirillov2023segany,zhang2023faster,xiong2023efficientsam}, the segmentation quality is measured by NoC@$90$/$95$ (average number of clicks required to achieve $90\%$ or $95\%$ IoU within a $20$-click budget) and $5$-mIoU (mIoU after five clicks) in high-precision IS tasks~\cite{liu2024rethinking,huang2024hrsam}. 
For efficiency evaluation on CPU devices, we employ two metrics: $20$-SPC (Seconds Per Click)~\cite{huang2024hrsam} and Online SPC. The $20$-SPC metric measures the average time to complete $20$ clicks, including preprocessing overhead (\eg, image encoding in SAM-based methods). In contrast, Online SPC captures the per-click latency during interaction, excluding preprocessing time. For traditional IS models preceding SAM, these two metrics yield identical values since they do not require preprocessing steps.

\subsection{Main Results}
\label{subsec:mainresult}

As shown in Table~\ref{tab:quantitative_comparison}, Inter2Former achieves SOTA performance across all evaluation metrics while maintaining competitive efficiency. 
The superior performance can be attributed to the dense-token design in Inter2Former's decoder. Unlike HRSAM++ which inherits SAM's sparse-token decoder and struggles with limited training data, Inter2Former's dense-token design enables more effective training from scratch, particularly beneficial when fine-tuning on high-quality datasets. In terms of efficiency, while Inter2Former shows marginally higher latency compared to HRSAM++, it maintains substantial speed advantages over other high-precision methods like SegNext while achieving better accuracy. Furthermore, Inter2Former demonstrates competitive efficiency against fast SAM variants like EfficientSAM~\cite{xiong2023efficientsam}, and significantly reduces the inference time compared to InterFormer~\cite{huang2023interformer}.


\begin{figure*}[t]
    \centering
    \begin{subfigure}[b]{0.49\linewidth}
        \centering
        \includegraphics[width=\linewidth]{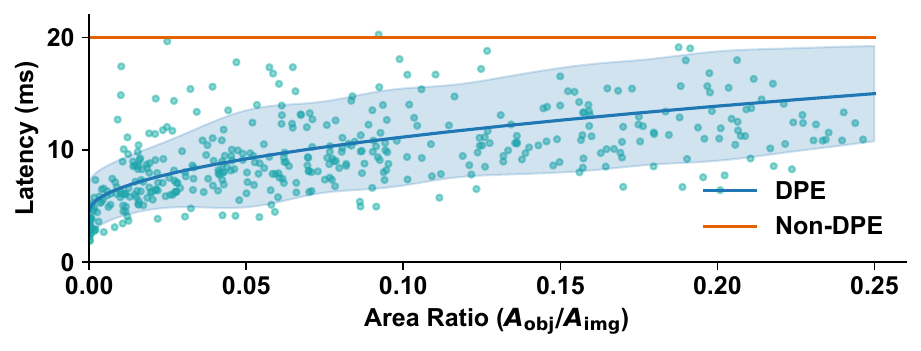}
        \caption{\textbf{DPE Efficiency}}
        \label{fig:dpe_eff}
    \end{subfigure}
    \hfill
    \begin{subfigure}[b]{0.49\linewidth}
        \centering
        \includegraphics[width=\linewidth]{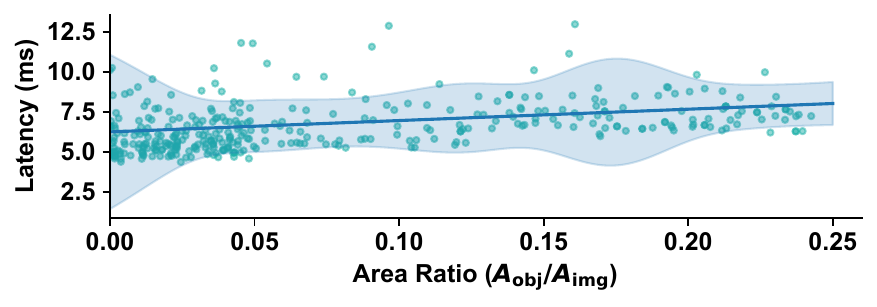}
        \caption{\textbf{DHA Efficiency}}
        \label{fig:dha_eff}
    \end{subfigure}
    
    \begin{subfigure}[b]{0.49\linewidth}
        \centering
        \includegraphics[width=\linewidth]{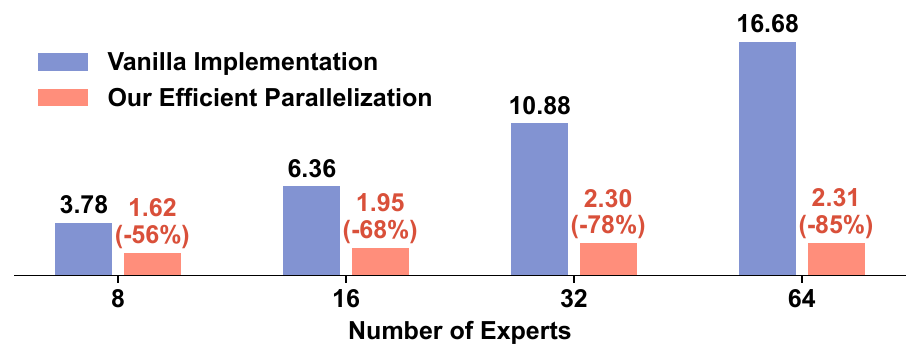}
        \caption{\textbf{HMoE Efficiency}}
        \label{fig:hmoe_eff}
    \end{subfigure}
    \hfill
    \begin{subfigure}[b]{0.49\linewidth}
        \centering
        \includegraphics[width=\linewidth]{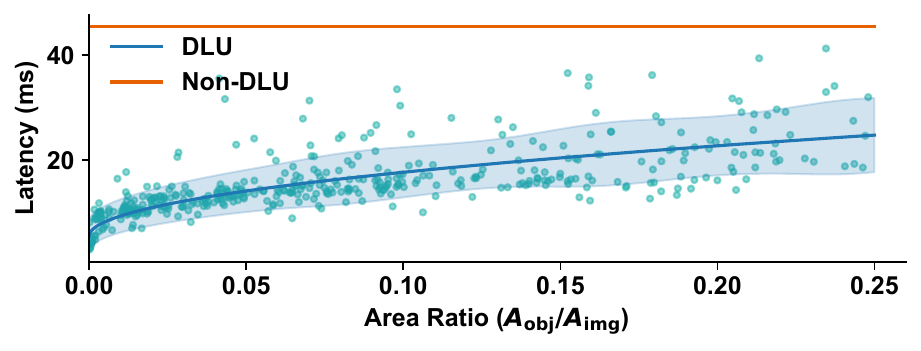}
        \caption{\textbf{DLU Efficiency}}
        \label{fig:dlu_eff}
    \end{subfigure}
    \vspace{-0.5em}
    \caption{Efficiency analysis. (a) Latency of DPE and Non-DPE (full prompt embedding). (b) DHA latency across different area ratios. (c) HMoE speedup with various expert numbers. (d) Latency of DLU and Non-DLU (full upsampling).}
    \label{fig:efficiency}
\end{figure*}

\begin{figure*}[ht]
  \centering
  \includegraphics[width=1.95\columnwidth]{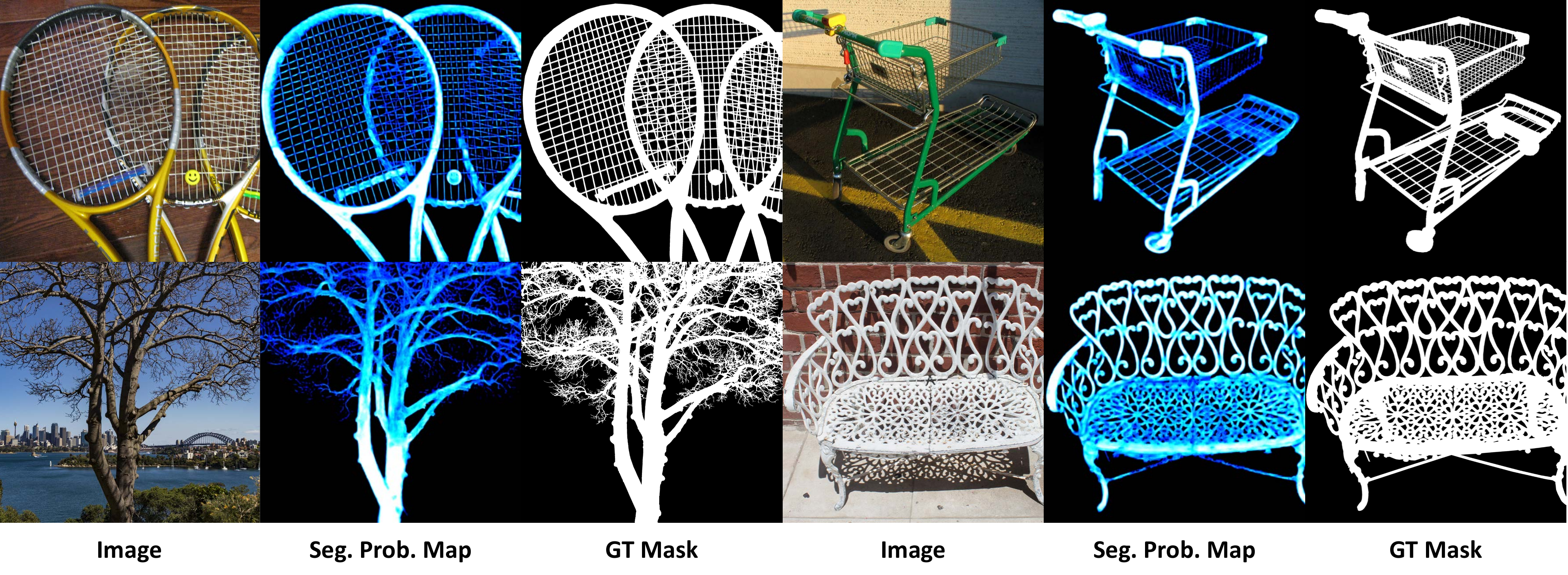}
  \vspace{-0.5em}
  \caption{Qualitative results. Inter2Former produces precise segmentation maps for challenging line structures under 20-click prompts.}
  \label{fig:qualitativeresults}
\end{figure*}

\subsection{Efficiency Analysis}
\label{subsec:efficiency}

We evaluate the computational efficiency of our proposed DPE, DHA and DLU on COCO dataset using object mask annotations. The diverse object scales provide objective benchmarks for efficiency analysis under various area ratios ($A_{\text{obj}}/A_{\text{img}}$). Additionally, we analyze HMoE's efficiency across different numbers of experts.

\noindent\textbf{DPE Efficiency.} As shown in Figure~\ref{fig:efficiency} (a), Non-DPE (with identical convolutional architecture) maintains a constant latency regardless of object size, while our DPE shows a linear increase with area ratios but maintains significant efficiency advantages. Notably, for predominant small objects, DPE requires $<25\%$ of the Non-DPE latency.

\noindent\textbf{DHA Efficiency.} As shown in Figure~\ref{fig:efficiency} (b), DHA shows a slow linear rise in latency as area ratios grow. This stable trend comes from DHA's focus on edges. While object area grows quadratically, the number of edge pixels grows more slowly, leading to efficient processing even for large objects.

\noindent\textbf{HMoE Efficiency.} As shown in Figure~\ref{fig:efficiency} (c), we compare our efficient parallel HMoE with vanilla implementation across expert counts. HMoE shows clear speedup with more experts, reaching 85\% latency reduction at 64 experts.

\noindent\textbf{DLU Efficiency.} As shown in Figure~\ref{fig:efficiency} (d), DLU as the inverse of DPE shows similar speed patterns. Our DLU runs much faster than Non-DLU with the same structure.

Detailed ablation studies for each module's performance are provided in the \textbf{supplementary material}.

\subsection{Qualitative Results}
\label{subsec:qualitativeresults}

As shown in Figure~\ref{fig:qualitativeresults}, Inter2Former achieves precise segmentation on challenging thin structures under 20 clicks.
\section{Conclusion}
This paper introduces Inter2Former for high-precision interactive segmentation, addressing the critical trade-off between performance and efficiency in dense-prompt-token processing. Through adaptive computation allocation implemented by the proposed Dynamic Prompt Embedding, Dynamic Hybrid Attention, Hybrid Mixture of Experts and Dynamic Local Upsampling, Inter2Former optimizes performance while maintaining efficiency, achieving SOTA performance with competitive speed on CPU devices.

\section*{Acknowledgments}
This work was supported by  the National Science Fund for Distinguished Young Scholars (No.62025603), the National Natural Science Foundation of China (No. U21B2037, No. U22B2051, No. U23A20383, No. 62176222, No. 62176223, No. 62176226, No. 62072386, No. 62072387, No. 62072389, No. 62002305 and No. 62272401), and the Natural Science Foundation of Fujian Province of China (No. 2021J06003, No.2022J06001).

{
    \small
    \bibliographystyle{ieeenat_fullname}
    \bibliography{main}
}
\clearpage
\onecolumn
\appendix

\section{Ablation Study}

To validate the effectiveness of each key component in Inter2Former, we conduct extensive ablation studies. We train different architectural variants from scratch for 80k iterations on HQSeg44K~\cite{sam_hq}, using HRSAM++ encoder~\cite{huang2024hrsam} distilled from SAM as the unified image encoder.

As shown in Table~\ref{tab:ablation}, replacing DPE with full prompt embedding (Non-DPE) shows marginal impact on model performance. In terms of attention mechanisms, our proposed hybrid attention design (DHA) achieves comparable results with Full Attention (All FA), while significantly outperforming pure BSQ Attention (All BSQA). For the upsampling module, although DLU shows slightly lower performance than full upsampling at 1024$^2$ resolution on DAVIS, this gap becomes negligible at 2048$^2$ resolution. These results demonstrate the effectiveness of our proposed lightweight modules.

Additionally, we compare our BSQA with VQA~\cite{lingle2023transformer} by replacing BSQA in our model, as shown in the last row of each input resolution in Table~\ref{tab:ablation}. The results indicate that VQA leads to performance degradation due to convergence issues during training, which validates the effectiveness of our BSQA design.

\begin{table}[!h]
  \centering
  \resizebox{\textwidth}{!}{
\begin{tabu}[c]{l c c c c c c c}
    \toprule
    \multirow{2}{*}{\textbf{Configuration}} & 
    \multirow{2}{*}{\textbf{Input Image Size}} &
    \multicolumn{3}{c}{\textbf{HQSeg44K} {\small$_{\textbf{Max H/W} > 4000}$}} & 
    \multicolumn{3}{c}{\textbf{DAVIS} {\small$_{\textbf{Max H/W} < 1000}$}} \\
    \cmidrule(lr){3-5} \cmidrule(lr){6-8}
    & & 5-mIoU $\uparrow$ & NoC90 $\downarrow$ & NoC95 $\downarrow$ &
    5-mIoU $\uparrow$ & NoC90 $\downarrow$ & NoC95 $\downarrow$ \\
    \midrule
    Inter2Former-Base & $1024\times1024$ 
    & 91.32 & 5.46 & 9.34 & 90.36 & 4.96 & 12.72 \\
    DPE $\rightarrow$ Non-DPE & $1024\times1024$ 
    & 91.33 & 5.44 & 9.38 & 90.56 & 4.94 & 12.48 \\
    DHA $\rightarrow$ All FA & $1024\times1024$ 
    & 91.37 & 5.38 & 9.31 & 90.35 & 5.01 & 12.24 \\
    DHA $\rightarrow$ All BSQA & $1024\times1024$ 
    & 89.71 & 6.19 & 10.12 & 88.20 & 5.83 & 13.53 \\
    DLU $\rightarrow$ Non-DLU & $1024\times1024$ 
    & 91.52 & 5.35 & 9.16 & 90.90 & 4.89 & 11.81 \\
    BSQA $\rightarrow$ VQA & $1024\times1024$ 
    & 90.91 & 5.59 & 9.58 & 89.59 & 5.31 & 12.75 \\
    \midrule
    Inter2Former-Base & $2048\times2048$ 
    & 92.68 & 4.24 & 7.39 & 92.00 & 4.29 & 7.82 \\
    DPE $\rightarrow$ Non-DPE & $2048\times2048$ 
    & 92.86 & 4.19 & 7.33 & 92.17 & 4.28 & 7.94 \\
    DHA $\rightarrow$ All FA & $2048\times2048$ 
    & 92.61 & 4.24 & 7.39 & 92.26 & 4.20 & 7.78 \\
    DHA $\rightarrow$ All BSQA & $2048\times2048$ 
    & 90.12 & 5.64 & 8.80 & 89.31 & 5.37 & 9.75 \\
    DLU $\rightarrow$ Non-DLU & $2048\times2048$ 
    & 92.76 & 4.22 & 7.32 & 92.13 & 4.30 & 7.90 \\
    BSQA $\rightarrow$ VQA & $2048\times2048$ 
    & 91.07 & 4.82 & 8.01    & 90.31 & 4.73 & 8.86 \\
    \bottomrule
\end{tabu}
}
  \vspace{-0.5em}
  \caption{
  Ablation study. Inter2Former-Base represents our complete model configuration. The arrow ($\rightarrow$) indicates replacement of our proposed module with alternatives: Non-DPE replaces DPE with full prompt embedding, All FA and All BSQA replace DHA with Full Attention and BSQ Attention respectively, Non-DLU substitutes DLU with full upsampling module, 
  and VQA replaces BSQA in DHA.
}
  \label{tab:ablation}
\end{table}

\begin{figure}[!h]
    \centering
    \includegraphics[width=0.8\columnwidth]{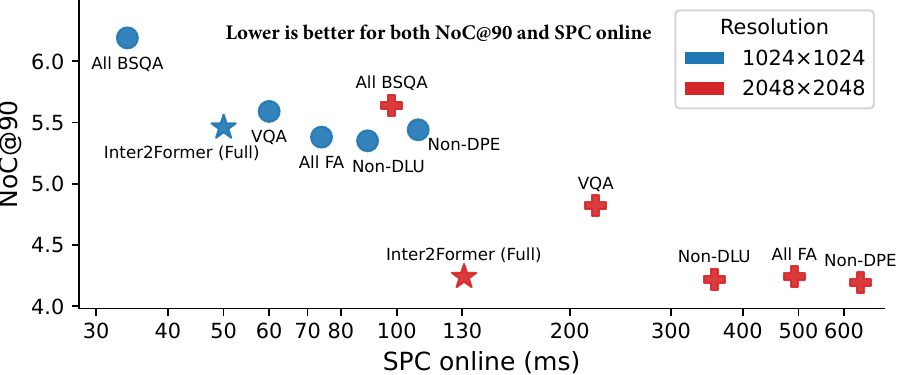}
    \caption{Performance-efficiency trade-off in the ablation study. We plot segmentation quality (NoC@90) against online inference latency (SPC online) for the architectural variants. The comparison, shown for both 1024×1024 and 2048×2048 resolutions, visually demonstrates that the full Inter2Former configuration achieves a superior balance of performance and efficiency compared to the ablated versions.}
    \label{fig:supp_ablation_v2}
\end{figure}

\end{document}